\documentclass[10pt,twocolumn,letterpaper]{article}

\usepackage[pagenumbers]{cvpr} 

\usepackage{graphicx}
\usepackage{amsmath,amssymb}
\usepackage{booktabs}

\usepackage[pagebackref,breaklinks,colorlinks]{hyperref}

\usepackage[capitalize]{cleveref}
\crefname{section}{Sec.}{Secs.}
\Crefname{section}{Section}{Sections}
\Crefname{table}{Table}{Tables}
\crefname{table}{Tab.}{Tabs.}

\def\paperID{} 
\def\confName{CVPR}
\def\confYear{2026}

\title{Binary Verification for Zero-Shot Vision}

\author{Rongbin Hu\\
mycube.tv\\
{\tt\small rongbinhu@mycube.tv}
\and
Jeffrey Liu\\
mycube.tv\\
{\tt\small jeffreyliu@mycube.tv}
}

\begin{document}
\maketitle


\begin{abstract}
We propose a training-free, \textbf{binary verification} workflow for zero-shot vision with off-the-shelf VLMs. It comprises two steps: ($i$) \emph{quantization}, which turns the open-ended query into a multiple-choice question (MCQ) with a small, explicit list of unambiguous candidates; and ($ii$) \emph{binarization}, which asks one \texttt{True/False} question per candidate and resolves deterministically: if exactly one is \texttt{True}, select it; otherwise, revert to an MCQ over the remaining plausible candidates. We evaluate the workflow on referring expression grounding (REC), spatial reasoning (Spatial-Map, Spatial-Grid, Spatial-Maze), and BLINK-Jigsaw. Relative to answering open-ended queries directly, \emph{quantization to MCQ} yields large gains, and \emph{True/False binarization} provides a consistent additional boost. Across all tasks, the same workflow produces significant improvements, indicating generality. We further integrate the proposed REC workflow into a real-world video processing and editing system, and present the system architecture and end-to-end pipeline in the paper. Together, these components define a unified inference-time workflow that offers a practical, drop-in path to stronger zero-shot vision with today’s VLMs.

\end{abstract}

\section{Introduction}

Zero-shot vision~\cite{frome2013devise, xian2019zsl} is the ability of a model to perform a new visual task without any task-specific training or fine-tuning. It is increasingly critical for science and deployment. Compared with supervised learning, which requires task-specific annotations that are often expensive and time-consuming to build, zero-shot vision enables rapid and cost-effective adaptation to new tasks and domains. Moreover, proprietary vision language models (VLMs) often deliver stronger out-of-the-box performance and ship with mature APIs and infrastructure that simplify integration, but they typically restrict task-specific training beyond prompting, making inference-time methods especially valuable.

Modern VLMs~\cite{openai2024gpt4o, liu2023llava, wang2024cogvlm, bai2023qwenvl,bao2022beit, dai2023instructblip, he2022mae, hong2024cogagent, liu2024llavaNext} show strong zero-shot ability on tasks such as image captioning, broad scene recognition, and short-form visual QA. In typical zero-shot use, the model answers an \emph{open-ended} prompt, for example, ``\emph{Describe this image}''. However, for more demanding tasks such as precise grounding, spatial reasoning, or fine-grained recognition, performance is often unreliable without task-specific fine-tuning. This unreliability stems from the interaction of several factors: free-form outputs make small numeric and referential errors both frequent and difficult to detect; model confidences are poorly calibrated, undermining decision thresholds; and more.

A fundamental idea in computing is that any number can be represented in a binary format. Quantizing and binarizing values to bits simplify arithmetic for machines and enable scalable computation. We adopt an analogous view and instantiate the idea of \emph{quantization} and \emph{binarization} for zero-shot vision tasks with a simple, training-free workflow. First, we \emph{quantize} the task by generating candidate units, e.g. detector boxes, grid cells, or entity pairs, so an open query becomes a multiple choice question (MCQ) with a small set of explicit hypotheses. Next, we \emph{binarize} the decision by asking, for each candidate $h_i$, a single yes/no claim that checks whether $h_i$ matches the description, and the VLM must answer either \texttt{True} or \texttt{False}.  Finally, we resolve outcomes deterministically: if exactly one candidate is \texttt{True}, we return it; if multiple answers are \texttt{True}, we restrict to that subset and issue a compact MCQ; if all are \texttt{False}, we fall back to the original MCQ. 

We evaluate the unified workflow across three task families using the same VLMs without any task-specific fine-tuning, and observe significant improvements on every task. 
\begin{itemize}
    \item \textbf{Referring Expression Comprehension (REC).} REC~\cite{qiao2020rec_survey} is the task of locating the specific object described by a natural-language expression in a given image. On REC (RefCOCO/+/g)~\cite{yu2016refcoco, mao2016refcocog}, our binary verification workflow with GPT-4o reaches ACC@0.5 of 71.7\%, 67.1\%, and 64.8\%, respectively, substantially above zero-shot GroundingDINO~\cite{liu2023groundingdino} (50.4\%, 51.4\%, 60.4\%). Note that the original GPT-4o only has ACC@0.5 of 8.4\%, 7.3\%, 9.1\%, respectively. 

   \item \textbf{Spatial reasoning (Map, Grid, Maze).} For spatial reasoning~\cite{wang2024spatial}, we evaluate three tasks, namely, Spatial-Map, Spatial-Grid, and Spatial-Maze, which probe directional relations, grid area recognition, and path finding, respectively. We take the original task interface as the baseline and then apply our two-step binary verification workflow. On \textbf{Spatial-Map}, the baseline is already MCQ. Binarization improves over MCQ from 78.3\% to 84.5\% and from 25.5\% to 31.3\% with GPT-4o~\cite{openai2024gpt4o} and CogVLM~\cite{wang2024cogvlm}, respectively. On \textbf{Spatial-Grid}, spatial quantization, i.e., overlaying explicit grid on the original image, boosts the classification accuracy from 47.2\% to 95.0\% and from 23.2\% to 58.8\% with GPT-4o and CogVLM, respectively. On \textbf{Spatial-Maze}, spatial quantization by overlaying an explicit grid is again essential, which improves the accuracy of finding the right path from 16.5\% to 68.7\%. An embedding local T/F corridor checks further improves the accuracy to 75.7\%. 

    \item \textbf{BLINK-Jigsaw.} BLINK~\cite{fu2024blinkmultimodallargelanguage} is a multiple-choice benchmark that reformats classic vision tasks into image prompts to probe core perceptual skills that humans solve “in a blink,” but that current VLMs often miss. To test whether our approach helps on these harder perception cases, we evaluate a modified and harder BLINK-Jigsaw setting. Our method improves accuracy from 51.6\% (MCQ) to 56.8\%, demonstrating that binary verification still delivers measurable gains even when the underlying task is challenging for state-of-the-art VLMs.

\end{itemize}

This work is driven by a need in our video product. Live streaming and video recording of many events often provide only a single wide camera feed, yet viewers expect a multi-camera and professionally directed stream. Our system re-organizes a single-camera video into a structured layout by identifying key visual components, enabling virtual camera movement and composited views. It must distinguish the active speaker from the audience in a seminar/demo day, track the in-play ball versus other balls in the scene in basketball, and identify the bride in a wedding. This motivates our focus on reliable grounding and is why we start from REC as a benchmark and prioritize accuracy as a first-order requirement. We present the end-to-end video processing and editing system, and show how the proposed binary verification workflow integrates into its architecture in live-streaming settings.

This paper makes the following contributions:
\begin{itemize}
    \item \textbf{Method.} We introduce a simple and unified workflow that \emph{quantizes} an open-ended query to a small hypothesis set for a MCQ and then \emph{binarizes} them into per-candidate True/False checks with deterministic resolution.
    \item \textbf{Experiment.} Using the same/similar workflow, we show broad gains across five tasks in three families, where quantization improves over open-ended prompting and binarization provides a further consistent boost.
    \item \textbf{System.} We present an end-to-end video processing and editing system (in live-streaming settings) that integrates the workflow for REC, turning grounded targets into downstream actions such as virtual camera movement, composited views, and selective annotation/redaction.
\end{itemize}

\section{Related Work}
\label{sec:related}

\noindent\textbf{Inference-time workflows.}
Modern LLM/VLM usage increasingly treats generation as a \emph{procedure} rather than a single shot. Chain-of-thought~\cite{wei2022chain} elicits intermediate reasoning, ReAct~\cite{yao2022react} interleaves tool use with reasoning, and modular frameworks (e.g., DSPy~\cite{khattab2024dspy}) compose these behaviors declaratively. Our work adopts this workflow view for vision, but focuses on a verification-centric interface.

\noindent\textbf{Verification vs.\ selection.}
A large body of methods improves answers by \emph{selection} over multiple candidates, including self-consistency/majority voting~\cite{wang2023selfconsistency}, process-of-elimination prompting~\cite{zhou2023poe}, LLM-as-a-judge~\cite{liu2023geval, zheng2023scoring}. These approaches remain MCQ-style, typically relying on sampling or confidence margins. In contrast, we convert each option into an independent \texttt{True}/\texttt{False} claim and apply a deterministic, codeword-style resolution over the boolean pattern.

\noindent\textbf{Quantization via proposals.}
Zero-shot recognition and localization leverage vision–language pretraining (CLIP~\cite{radford2021clip}, ALIGN~\cite{jia2021align}) for open-vocabulary categorization, extend to localization with text-conditioned detectors (GLIP~\cite{li2022glip}, OWL-ViT~\cite{minderer2022owlvit}, GroundingDINO~\cite{liu2023groundingdino}), and expose generic masks via foundation segmentation models (SAM)~\cite{kirillov2023sam}. We streamline zero-shot vision by a unified workflow. 

\noindent\textbf{Spatial reasoning benchmarks.}
Recent work shows that VLMs struggle on compositional and geometric reasoning even under MCQ interfaces. We adopt the NeurIPS 2024 Spatial-Reasoning suite (Map, Grid, Maze)~\cite{wang2024spatial} as a benchmark substrate and demonstrate that explicit spatial quantization by visible grids and indices, combined with binarization, substantially reduces error across all three tasks.

\noindent\textbf{Our position.}
Relative to sampling-heavy selection and program-heavy tool pipelines, we propose a simple, unified, training-free workflow to streamline zero-shot vision tasks. 

\section{Methodology}
\label{sec:method}

Our method reframes prediction as verification through two transformations: \emph{quantization} and \emph{binarization}. Quantization converts an open query into a MCQ by proposing a compact, explicit set of hypotheses. Binarization converts that MCQ into a set of yes/no tests by rewriting each hypothesis as a precise claim. A VLM answers only \texttt{True} or \texttt{False}. Decisions are made by deterministic resolution over these binary outcomes.

\subsection{Quantization}
We first quantize the original task into a $K$-way discrete alphabet $\mathcal{A}=\{a_1,\dots,a_K\}$. For a given input, we then form a shortlist $S\subseteq\mathcal{A}$ with $|S|=m$ by proposing hypotheses that are \emph{concrete and checkable}. Each hypothesis must be unambiguously referable in language (e.g., via coordinates, identifiers, or canonical labels). The design goal is \emph{high coverage with parsimony}: include the true state with high probability while keeping $|S|$ small enough to verify efficiently. After this step, the problem is an MCQ over the finite state space $S$.

In practice, quantization is \emph{task-specific}. We quantize either the \emph{answer space} (i.e., the admissible outputs) or the \emph{search space} from which answers are derived. Concretely,  (i) for object- or person-centric queries, use an open-vocabulary detector or tracker to produce a shortlist of regions/tracks with bounding boxes, masks, and/or IDs; (ii) for spatial queries, overlay an indexed grid to discretize the image plane, treating cells as candidates; (iii) for tasks with native discrete labels (e.g., expressions, attributes, compass directions), adopt the given label set as the quantized alphabet; (iv) when necessary, prompt a VLM/LLM to enumerate plausible answers under task-specific constraints. Overall, any proposal mechanism is admissible if it yields hypotheses that can be stated as precise claims about the input. 

\subsection{Binarization}
\label{sec:method_binarization}

Binarization replaces multi–way choice with a fixed set of independent True/False questions, one per MCQ option. For a $m$–option MCQ, we ask $m$ separate questions of the form ``\emph{Is option $i$ the correct answer?}'' and require the VLM to return \texttt{True} or \texttt{False} to each question. The $m$ answers form a boolean pattern over the option set, and the decision is made by deterministic rules that operate on this pattern:
\begin{itemize}
    \item\textbf{Single-\texttt{True}.} If exactly one option is \texttt{True}, return it immediately.
    \item\textbf{Multiple-\texttt{True}.} If more than one option is \texttt{True}, treat this as verification noise and a useful constraint. Reduce the MCQ to the subset marked \texttt{True} and re-run per-option True/False on this reduced set. If the maximal allowed retries is reached without isolating a single \texttt{True}, construct a final MCQ containing only the \texttt{True} options and issue a single-shot MCQ prompt, and take that single-shot choice as the answer.
    \item\textbf{All-\texttt{False}.} If all options are \texttt{False}, regard this as under–specification. If retries remain, re-run binarization. If the maximal retries are reached, construct a final MCQ over all options and answer it in a single shot. 
\end{itemize}

Concretely, for four options $(a,b,c,d)$, the pattern \texttt{[True, False, False, False]} selects $a$ immediately, whereas \texttt{[True, False, True, False]} shrinks the MCQ to $\{a,c\}$ and triggers a refined two–way round. The deterministic resolution borrows the idea of \emph{error detection} from channel coding~\cite{hamming1950error}: per-option True/False outcomes act as a short codeword, enabling simple recovery without relying on probability margins. Binarization is largely \emph{task-agnostic}. Any candidate set, however produced, is handled by the same per-option True/False queries with deterministic resolution.

Binary verification exposes a simple ``\emph{certainty knob}'': instruct the verifier to treat uncertainty as \texttt{False}. This reduces false positives for Single-\texttt{True} and sharpens pruning when Multiple-\texttt{True} occurs. At the extreme, if the \texttt{True} criterion is so strict that every round is All-\texttt{False}, the scheme degenerates to the original single-shot MCQ. Hence the strictness of “\texttt{True}” can be the tuning parameter: looser thresholds may lead to a larger potential gain, while tighter thresholds favor a higher chance of non-negative gain.

\subsection{Performance, Efficiency, Complexity}

\noindent\textbf{Performance.}
Binary verification improves accuracy through three mechanisms. First, quantization shrinks the search space by mapping an open-ended query to a small, explicit candidate set. Second, binarization reduces cross-candidate interference by localizing each decision to a single highlighted unit, limiting distraction when many proposals are present. Third, the deterministic resolution in binarization prunes the hypothesis set whenever single \texttt{True} or multiple \texttt{True} labels occur, improving the probability of a correct final choice. 

\vspace{4pt}\noindent\textbf{Efficiency.}
Quantization controls cost by shrinking the MCQ shortlist, yielding a predictable budget that scales with shortlist size. In contrast, binarization latency is governed by the number of verification rounds (RTTs), not by the number of options, since per-option \texttt{True/False} checks can be issued in parallel. With one round, binary verification matches MCQ latency in the Single-True case; in the Multiple-True and All-False cases, it requires one extra round (one additional RTT). 

\vspace{4pt}\noindent\textbf{Complexity.}
We intentionally keep the resolution rule simple and practical. The T/F binarization already adds calls and therefore latency and cost.  Piling on soft aggregation or heavier fusion schemes (e.g., noisy-OR~\cite{heckerman1993causalindependence} and BP/CRF pairwise consistency~\cite{krahenbuhl2011efficient}) would further inflate compute and reduce reproducibility. In our deployment-oriented zero-shot setting, a deterministic, parameter-free rule is stable and compute-fair, while score-based or iterative alternatives add complexity for little gain.


\section{Experiment}
\label{sec:experiment}

We study five vision tasks from three diverse families, namely REC~\cite{qiao2020rec_survey}, Spatial-Map, Spatial-Grid, Spatial-Maze, in spatial reasoning family~\cite{wang2024spatial}, and BLINK-Jigsaw~\cite{fu2024blinkmultimodallargelanguage}, with the single binary verification workflow. We evaluate the workflow mainly with a leading proprietary VLM (GPT-4o~\cite{openai2024gpt4o}) and a strong open-source VLM (CogVLM~\cite{wang2024cogvlm}), each representative of the state of the art in its category. In our experiments, we disable retries, where after quantization, we make a single pass of per-option True/False queries. If the outcome is not Single-\texttt{True}, we conclude with one final single-shot MCQ over the \texttt{True} subset if any, otherwise over all options.

\subsection{Referring Expression Grounding (REC)}
\label{sec:exp-rec}

\begin{figure*}[!t]
    \centering
    \includegraphics[width=\textwidth]{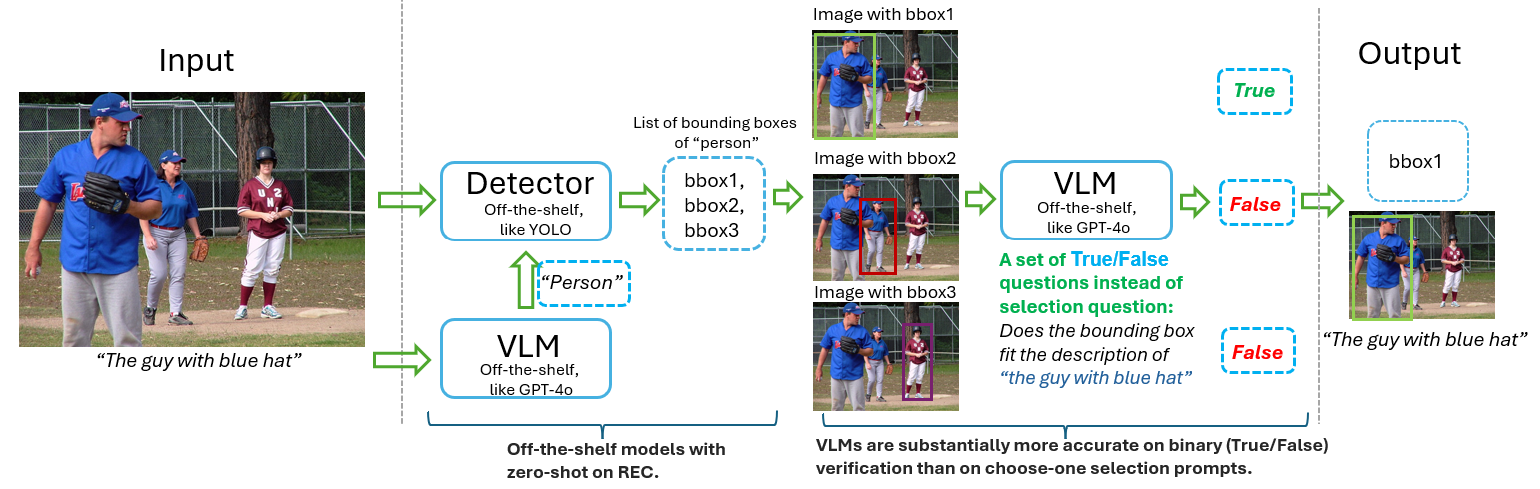}
    \caption{An example of binary verification workflow for REC.}
    \label{fig:rec_workflow}
\end{figure*}

\begin{table*}[t]
\centering
\footnotesize
\setlength{\tabcolsep}{3pt}
\renewcommand{\arraystretch}{0.95}
\caption{REC benchmarking (ACC@0.5, \%).}
\label{tab:bench}
\vspace{-4pt}
\resizebox{0.95\textwidth}{!}{%
\begin{tabular}{l l ccc ccc cc}
\hline
Regime & Method &
\multicolumn{3}{c}{RefCOCO} &
\multicolumn{3}{c}{RefCOCO+} &
\multicolumn{2}{c}{RefCOCOg} \\
\cline{3-10}
& & Val & TestA & TestB & Val & TestA & TestB & Val & Test \\
\hline
Supervised                    & CogVLM (REC-trained)                              & 92.6 & 94.3 & 91.5 & 85.2 & 89.6 & 79.8 & 88.7 & 89.4 \\
Supervised                    & GroundingDINO (REC-trained)                       & --   & 77.3 & 72.5 & --   & 72.0 & 59.3 & --   & 66.3 \\
\hline
Workflow with REC             & GroundingDINO + CRG (REC-trained rerank)          & --   & 81.6 & 73.2 & --   & 77.0 & 60.0 & --   & 69.6 \\
\hline
Strict zero-shot              & GroundingDINO (zero-shot baseline)                & 50.4 & 57.2 & 43.2 & 51.4 & 57.6 & 45.8 & 60.4 & 59.5 \\
\hline
Zero-shot VLM (baseline)      & GPT-4o (organic / vanilla selection)              & 8.4  & --   & --   & 7.3  & --   & --   & 9.1  & --   \\
\hline
Zero-shot on REC        & MCQ (LLaVA, single-shot)                          & 34.7 & --   & --   & 34.6 & --   & --   & 44.4 & --   \\
Zero-shot on REC       & Binary verification (LLaVA)                       & 44.6 & --   & --   & 42.3 & --   & --   & 50.7 & --   \\
Zero-shot on REC       & MCQ (GPT-4o, single-shot)                         & 62.1 & 65.6 & 58.4 & 59.6 & 61.0 & 53.7 & 62.2 & 61.0 \\
Zero-shot on REC       & MCQ (GPT-4o, majority voting)                     & 63.6 & --   & --   & 60.2 & --   & --   & 62.6 & --   \\
\textbf{Zero-shot on REC (ours)} & \textbf{Binary verification (GPT-4o)}           & \textbf{71.7} & \textbf{77.8} & \textbf{63.0} & \textbf{67.1} & \textbf{70.1} & \textbf{58.5} & \textbf{64.8} & \textbf{63.7} \\
Zero-shot on REC (ours) & {Binary verification (GPT-5)}           & 79.3 & 85.6 & 70.4 & 74.2 & 80.4 & 65.2 & 72.4 & 71.5 \\
\hline
\end{tabular}%
}
\vspace{-6pt}
\end{table*}


\begin{table}[t]
\centering
\footnotesize
\setlength{\tabcolsep}{3pt}
\renewcommand{\arraystretch}{0.95}
\caption{RefCOCO (Val) binarization breakdown (GPT-4o).}
\label{tab:refcoco_breakdown}
\vspace{-3pt}
\begin{tabular}{lccc}
\toprule
 & \textbf{Single-\texttt{True}} & \textbf{Multiple-\texttt{True}} & \textbf{All-\texttt{False}} \\
\midrule
\textbf{Frequency (\%)} & 41.0 & 45.5 & 11.0 \\
\textbf{Candidate}      & $3.6\rightarrow 1$ & $7.0\rightarrow 4.2$ & $3.5\rightarrow 3.5$ \\
\textbf{ACC@0.5, \%}        & 80.8 & 71.7 & 54.8 \\
\textbf{ACC@0.5, \%, MCQ}  & 72.4 & 58.2 & 54.8 \\
\bottomrule
\end{tabular}
\vspace{-5pt}
\end{table}

\noindent\textbf{Task description.}
REC is the task of locating the specific object described by a natural language expression (e.g., `the red mug on the left') in a given image. The standard output is a single bounding box that identifies the referent. 

\vspace{4pt}\noindent\textbf{Quantization.}
As illustrated in Fig.~\ref{fig:rec_workflow}, from the expression, we prompt the VLM to name a coarse object class $c^\star$ (e.g., \textit{person}, \textit{cat}), then run a class-conditioned detector on the image for $c^\star$. We shortlist purely by detector confidence: retain only bounding boxes with score $\ge \tau$ and index them $\{a_1,\dots,a_m\}$, with $m$ varying per image. 

\vspace{4pt}\noindent\textbf{Binarization.}
For each shortlisted box $a_i$, we present the original image with bounding box $i$ overlaid and pose the same binary question: \emph{“Does the object in the highlighted bounding box match the referring description? Answer with True or False.”} The referring expression is appended verbatim. The VLM returns exactly \texttt{True} or \texttt{False}. After one pass over all boxes, we apply deterministic resolution: if exactly one box is \texttt{True}, we return it; if multiple boxes are \texttt{True}, we show only those boxes and issue a single-shot MCQ to select the best match; if all boxes are \texttt{False}, we issue a single-shot MCQ over the original shortlisted boxes.  

\vspace{4pt}\noindent\textbf{Results.}
We evaluate on RefCOCO, RefCOCO+, and RefCOCOg, i.e., the standard REC benchmarks derived from MS-COCO. A detection is correct if its Intersection over Union (IoU) with the ground-truth box exceeds 0.5. We therefore report ACC@0.5. 

We compare three supervised REC baselines, including \emph{CogVLM} (REC-trained VLM)~\cite{wang2024cogvlm}, \emph{GroundingDINO}~\cite{liu2023groundingdino} fine-tuned for REC, and a REC-trained workflow \emph{GroundingDINO+CRG} that re-ranks detector proposals. Zero-shot baselines include \emph{GroundingDINO} without REC finetuning and \emph{GPT-4o (vanilla REC)} that directly outputs a box from a single prompt. Our zero-shot variants all use the same YOLO-World~\cite{cheng2024yoloworld} proposals and differ in the VLMs and workflows. Note that YOLO-World is trained without COCO for dataset cleanliness. \emph{Binary verification (LLaVA\footnote{CogVLM includes supervised training for REC, so we instead use LLaVA\mbox{-}vicuna\mbox{-}13b as the representative open-source VLM.}/GPT-4o/GPT-5\footnote{GPT-5 is included because REC has direct, real-world applications, so reporting the strongest available VLM is practically informative.})}~\cite{liu2023llava,openai_gpt5_system_card_2025} performs per-box True/False judgments with the deterministic resolution, while \emph{MCQ (LLaVA/GPT-4o)} issues a single-shot MCQ over proposals. We also include \emph{MCQ (GPT-4o, majority voting)} which repeats the single-shot MCQ three times and takes the vote. We use the default confidence level of 0.25 of YOLO-World, resulting in 5.02, 5.03, and 2.45 candidates on average for RefCOCO/+/g(Val), respectively. 
We use temperature 0.2, while majority-vote runs use temperature 1.0, including the experiments in the following sections. 

As shown in Table~\ref{tab:bench}, our \emph{binary verification (GPT-4o)} surpasses the zero-shot GroundingDINO baseline by a large margin. With GPT-5, binary verification further exceeds supervised methods with the sole exception of fully supervised CogVLM. Binary verification consistently outperforms MCQ, including MCQ with majority voting. Table~\ref{tab:refcoco_breakdown} breaks down RefCOCO (Val) by boolean pattern, reporting ACC@0.5, frequency, and candidate counts pre/post pruning, showing where the gain comes from. The last row gives MCQ accuracy on the same items. 2.5\% of items lack a valid candidate box. Note that the “organic” GPT-4o baseline attains less than 10\% ACC@0.5 on these benchmarks, suggesting that it is not REC-trained.

\begin{figure*}[t]
  \centering
  \newlength{\panelw}\setlength{\panelw}{0.31\textwidth}
  \newlength{\panelh}\setlength{\panelh}{0.78\panelw}

  \begin{subfigure}[t]{\panelw}
    \centering
    \includegraphics[width=\linewidth,height=\panelh,keepaspectratio]{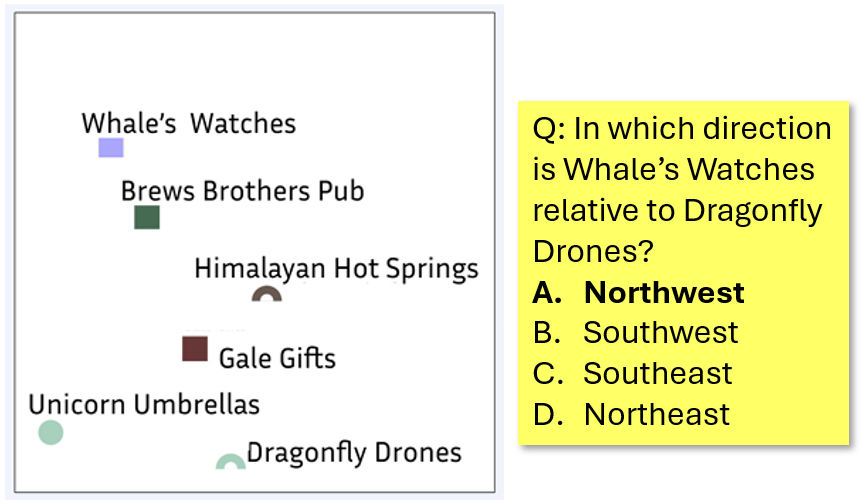}
    \subcaption{Spatial-Map example}
  \end{subfigure}\hfill
  \begin{subfigure}[t]{\panelw}
    \centering
    \includegraphics[width=\linewidth,height=\panelh,keepaspectratio]{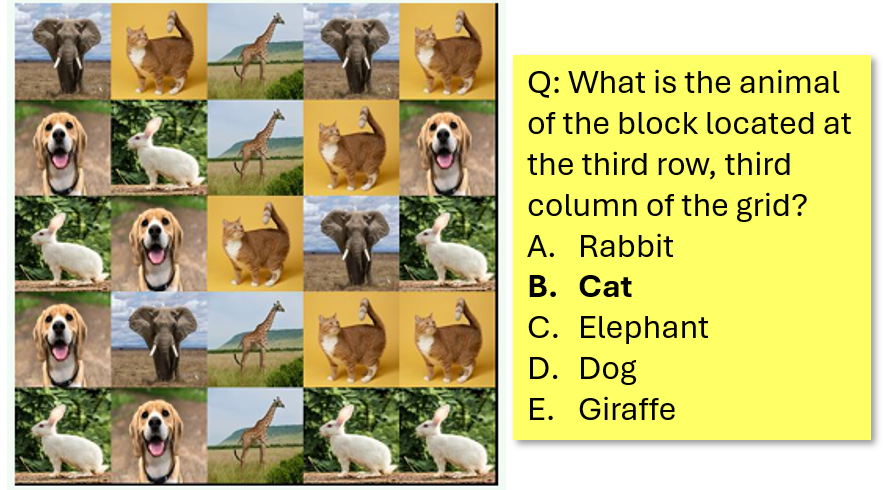}
    \subcaption{Spatial-Grid example}
  \end{subfigure}\hfill
  \begin{subfigure}[t]{\panelw}
    \centering
    \includegraphics[width=\linewidth,height=\panelh,keepaspectratio]{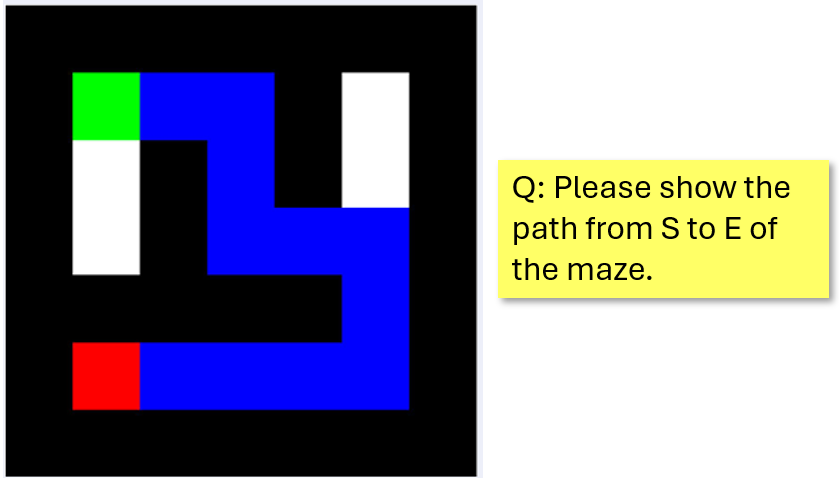}
    \subcaption{Spatial-Maze example}
  \end{subfigure}

  \caption{Illustrations for the Spatial-Reasoning family: for (a), we restrict to pairwise relations; for (b), we fix the query cell to (3,3), which is empirically harder; for (c), we directly request the full path (start, turns, end).}
  \label{fig:spatial_reasoning}
\end{figure*}

\subsection{Spatial-Map}
\label{sec:exp-spatial-map}

\noindent\textbf{Task description.}
The Spatial-Map~\cite{wang2024spatial} task presents a 2D map containing named locations and a natural-language question about a relative spatial relation between two targets, as shown in Fig.~\ref{fig:spatial_reasoning} (a). Spatial reasoning of this kind is known to be difficult for current VLMs due to sensitivity to absolute/relative spatial references.

\vspace{4pt}\noindent\textbf{Quantization.}
This task is natively an MCQ over a fixed set of directions, so we do not perform additional quantization. 

\vspace{4pt}\noindent\textbf{Binarization.}
Given a question “\emph{Where is \textit{A} relative to \textit{B}?}” with options $\{\text{NW},\text{SW},\text{SE},\text{NE}\}$, we turn each option into a True/False claim and query the VLM once per option. The input is the map and the fixed claim, e.g., “\emph{\textit{A} is northwest of \textit{B}. Answer: True or False.}” We issue the four binary queries individually and apply deterministic resolution. 

\vspace{4pt}\noindent\textbf{Results.}
We evaluate GPT-4o and CogVLM, under three inference interfaces: (i) \emph{MCQ} (single-shot choice among four directions), (ii) \emph{MCQ (5$\times$ MV)} which repeats MCQ five times with fixed prompt and returns the majority vote, and (iii) \emph{Binary verification}, our proposed binary True/False verification. Consistent with the REC findings, binary verification exceeds single-shot MCQ and MCQ majority voting. Table~\ref{tab:spatial_results} summarizes results.

\begin{table}[t]
\centering
\footnotesize
\setlength{\tabcolsep}{4pt}
\renewcommand{\arraystretch}{0.95}
\caption{Spatial-Map accuracy (\%).}
\label{tab:spatial_results}
\vspace{-3pt}
\begin{tabular}{lccc}
\hline
\textbf{VLM} & \textbf{MCQ} & \textbf{MCQ (5$\times$ MV)} & \textbf{Binary verification} \\
\hline
GPT-4o   & 78.3 & 78.7 & \textbf{84.5} \\
CogVLM  & 25.5 & 25.7 & \textbf{31.3} \\
\hline
\end{tabular}
\vspace{-5pt}
\end{table}

\subsection{Spatial-Grid}
\label{sec:exp-spatial-grid}

\noindent\textbf{Task description.}
The Spatial-Grid~\cite{wang2024spatial} task presents an image with an $5\times 5$ visually discrete lattice and a natural-language query that specifies a cell by row/column together with a multiple-choice list of categories (Fig.~\ref{fig:spatial_reasoning} (b)). The model must map the index to the correct grid cell and identify the object occupying that cell (e.g., \textit{cat}). This formulation isolates two capabilities: (i) positional reasoning over a visually discrete lattice and (ii) in-cell recognition.

\vspace{4pt}\noindent\textbf{Quantization.}
The grid in the source image is only \emph{visually} discrete, where it is likely that spatial reasoning conducted by a VLM is at continuous image level, rather than at cell level. We apply a \emph{spatial quantization}: overlay visible grid lines (e.g., $5{\times}5$) so that the image is partitioned into disjoint regions with unambiguous cell boundaries and indices, as shown in Fig.~\ref{fig:spatial_grid_quantization}. This creates a two-dimensional quantization: (i) a \emph{spatial alphabet} of cells defined by the overlay, and (ii) a \emph{semantic alphabet} of animal categories provided by the options. The spatial quantization converts a fuzzy, pixel-level notion of “third row, third column’’ into a stable, language-aligned coordinate frame with unambiguous cell boundaries and indices. Note that the grid and overlay can be determined and generated by the VLM itself for a given image, rather than pre-defined rules.

\vspace{4pt}\noindent\textbf{Binarization.}
We verify \emph{categories}, not cells. Given a query that specifies a target cell $(r,c)$ on the overlaid grid, we show the \emph{full} image with the grid and highlight $(r,c)$ and issue one True/False claim per candidate category $y$: “\emph{The object in cell $(r,c)$ is a \textbf{[y]}. Answer with True or False.}” The category list is the MCQ option set. After a single round over all categories, we apply the deterministic resolution.

\vspace{4pt}\noindent\textbf{Results.}
We evaluate GPT-4o and CogVLM on Spatial-Grid under five inference schemes: (1) \emph{MCQ (orig)} using the original image; (2) \emph{MCQ (orig, 5$\times$ MV)} repeating MCQ five times and taking the majority vote; (3) \emph{MCQ (grid)} using the overlaid grid image; (4) \emph{MCQ (grid, 5$\times$ MV)} with majority vote; and (5) \emph{Binary verification} using per-category True/False. As shown in Table~\ref{tab:spatialgrid_results}, spatial quantization dramatically improves performance. Binarization further boosts the performance.  

\begin{table*}[t]
\centering
\footnotesize
\setlength{\tabcolsep}{8pt}
\renewcommand{\arraystretch}{0.95}
\caption{Spatial-Grid accuracy (\%). MV = 5$\times$ majority vote.}
\label{tab:spatialgrid_results}
\vspace{-3pt}
\begin{tabular}{lccccc}
\hline
\textbf{VLM} & \textbf{MCQ (orig)} & \textbf{MCQ (orig, MV)} & \textbf{MCQ (grid)} & \textbf{MCQ (grid, MV)} & \textbf{Binary verif.} \\
\hline
GPT-4o  & 47.2 & 47.5 & 92.2 & 93.7 & \textbf{95.0}\\
CogVLM  & 23.2 & 20.4 & 52.1 & 55.2 & \textbf{58.8} \\
\hline
\end{tabular}
\vspace{-5pt}
\end{table*}

\subsection{Spatial-Maze}
\label{sec:exp-spatial-maze}

\noindent\textbf{Task description.}
The Spatial-Maze~\cite{wang2024spatial} task presents a 2D maze image with colored markers: the start cell \textbf{S} (green), the goal \textbf{E} (red), traversable corridor cells (blue) and walls (black). A natural language query asks for the path from \textbf{S} to \textbf{E} (Fig.~\ref{fig:spatial_reasoning} (c)). The path is reported in a set of critical point coordinates, including the starting point, all turning points, and the end point. Correctness is defined by the validity of the path. 

\begin{figure}[t]
  \centering
  \begin{subfigure}[t]{0.48\columnwidth}
    \centering
    \includegraphics[width=\linewidth,trim=4pt 4pt 4pt 4pt,clip]{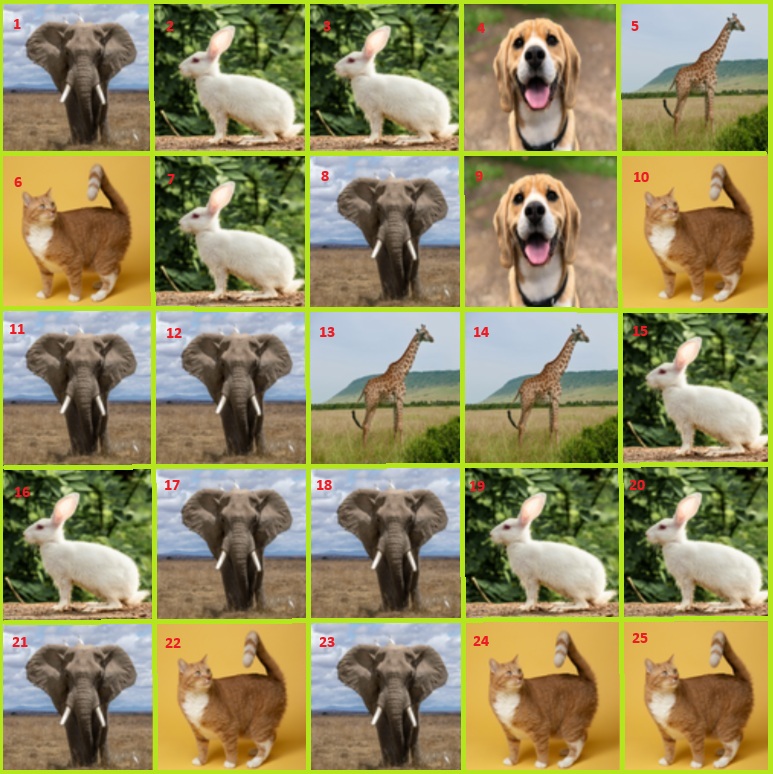}
    \caption{Spatial-Grid.}
    \label{fig:spatial_grid_quantization}
  \end{subfigure}
  \hfill
  \begin{subfigure}[t]{0.48\columnwidth}
    \centering
    \includegraphics[width=\linewidth,trim=4pt 4pt 4pt 4pt,clip]{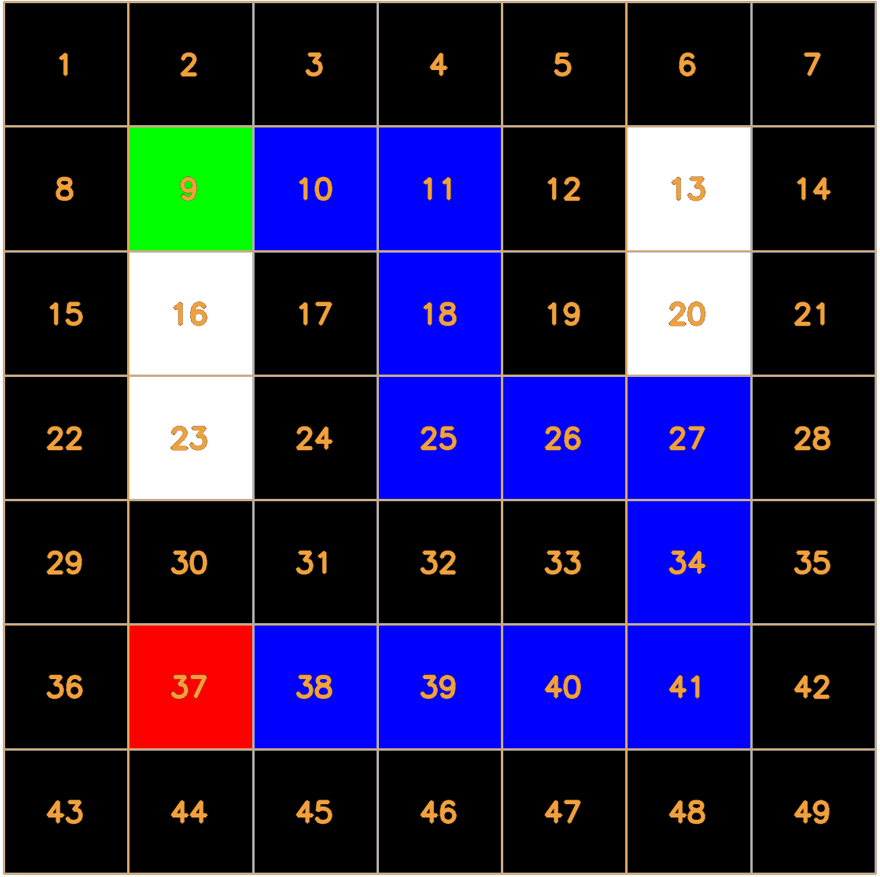}
    \caption{Spatial-Maze.}
    \label{fig:spatial_maze_quantization}
  \end{subfigure}
  \vspace{-4pt}
  \caption{Explicit spatial quantization via visible grid overlays.}
  \vspace{-6pt}
  \label{fig:spatial_quantization_both}
\end{figure}

\vspace{4pt}\noindent\textbf{Quantization.}
We apply spatial quantization by overlaying visible grid lines (e.g., $7{\times}7$) so the maze is partitioned into disjoint cells with boundaries and indices (Fig.~\ref{fig:spatial_grid_quantization}). The grid can be proposed by the VLM itself and rendered from its returned coordinates. Unlike Spatial-Grid, however, spatial quantization does not generate a small semantic alphabet: even with a fixed grid, the target output is a \emph{path}, and the induced $K$-way MCQ over all possible paths is combinatorial and potentially very large. Constructing a reliable shortlist that is both small and likely to contain the correct path is nearly equally difficult as constructing the right path.

\vspace{4pt}\noindent\textbf{Binarization.}
Spatial-Maze lacks a small MCQ shortlist, and we cannot apply the deterministic resolution on binary patterns proposed in Section~\ref{sec:method}. Instead, we embed binary verification inside the path-specification prompt. The model must output (i) an ordered list of indexes of the cells on the path, and (ii) a parallel list of per-step True/False checks of the form: “\emph{for each intermediate cell on the proposed path, is the cell blue}?”. This variant preserves the principle of performing simpler verifications. However, it does not yield the same option-wise binarization in MCQ settings and the deterministic resolution.

\vspace{4pt}\noindent\textbf{Results.}
We evaluate Spatial-Maze with GPT-4o and CogVLM. A prediction is correct only if the entire route is valid, verified either from the coordinate set (start, turns, end) or from the sequence of cell indices. CogVLM almost never returns a valid path, suggesting that this task exceeds its current capability. We therefore omit its results. For GPT-4o, we report three settings: (i) \emph{no grid} (original maze), (ii) \emph{with grid overlay}, and (iii) \emph{with grid + T/F prompting}. Accuracy improves with quantization dramatically and further with binary verification. 

\begin{table}[t]
\centering
\small
\setlength{\tabcolsep}{6pt}
\renewcommand{\arraystretch}{0.95}
\caption{Spatial-Maze accuracy (\%). }
\label{tab:maze_results}
\begin{tabular}{lc}
\hline
\textbf{Method (GPT-4o)}                 & \textbf{Accuracy} \\
\hline
No grid (original image)                 & 16.5 \\
Grid overlay                             & 68.7 \\
Grid overlay + T/F prompting       & \textbf{75.7} \\
\hline
\end{tabular}
\end{table}

\subsection{BLINK-Jigsaw}
\label{sec:exp-blink}

\begin{figure}[t]
  \centering
  \includegraphics[width=0.70\columnwidth]{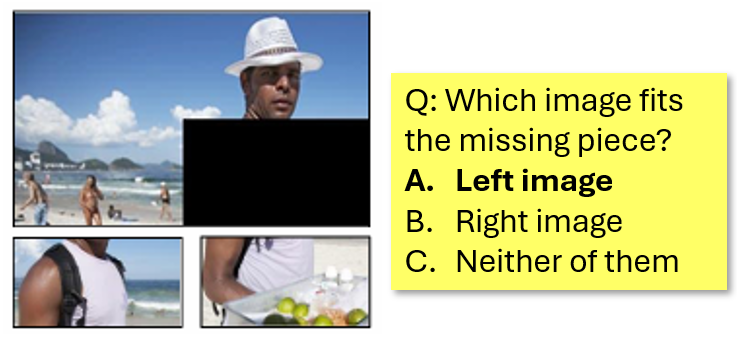}
  \caption{An example of BLINK-Jigsaw.}
  \label{fig:blink_jigsaw}
\end{figure}

\begin{table}[t]
\centering
\small
\setlength{\tabcolsep}{6pt}
\renewcommand{\arraystretch}{0.95}
\caption{BLINK-Jigsaw acc. (\%) with 95\% CIs (t).}

\label{tab:blink_results}
\begin{tabular}{lcc}
\hline
\textbf{Method (GPT-4o)} & \textbf{Acc.} & \textbf{95\% CI} \\
\hline
MCQ (Single-shot)                 & 51.6 & [49.6, 53.5] \\
MCQ (MV)                          & 45.2 & [43.9, 46.5] \\
Binary verification (Normal)      & 52.8 & [51.4, 54.2] \\
Binary verification (Certainty)   & \textbf{56.8} & [55.5, 58.0] \\
\hline
\end{tabular}
\end{table}

\noindent\textbf{Task description.}
BLINK~\cite{fu2024blinkmultimodallargelanguage} is a multiple-choice benchmark that reformulates 14 classic vision tasks (e.g., relative depth, correspondence) into single/multi-image prompts to probe core visual perception that humans solve “\emph{in a blink},” but current VLMs often miss. As shown in Fig.~\ref{fig:blink_jigsaw}, the BLINK–Jigsaw subtask is to select the tile that correctly completes the image. To better reflect real-world deployments where a quantized option set may not cover the true answer, we extend the original two-way setup to a three-way MCQ by adding option (C) \emph{Neither image is correct}. The “reject” choice also prevents a model from implicitly converting the task to a True/False decision (i.e., ``\emph{is A more likely than B to be the missing tile}'', rather than examining if A or B is indeed the missing tile. This increasing difficulty \emph{significantly}. Compared with other BLINK subtasks, Jigsaw is more purely visual, less reasoning-heavy, and overlaps less with our other experiments, making it a complementary stress test.

\vspace{4pt}\noindent\textbf{Quantization.}
The BLINK-Jigsaw is a three-way MCQ. 

\vspace{4pt}\noindent\textbf{Binarization.}
We follow the same binarization workflow by converting the MCQ into three True/False queries per-option, followed by the deterministic resolution. Because this task is especially challenging for current VLMs, we enable the certainty knob (Section~\ref{sec:method_binarization}) by instructing: “if uncertain, answer \texttt{False}” in each verification prompt. 

\vspace{4pt}\noindent\textbf{Results.}
We report results on GPT-4o under four inference modes: (i) \emph{MCQ}; (ii) \emph{MCQ (3$\times$ MV)}, majority vote with three repeats; (iii) \emph{Binary verification (normal)}, the True/False procedure without the `uncertain $\rightarrow$ \texttt{False}'' policy; and (iv) \emph{Binary verification (certainty policy)}, the True/False procedure with the ``uncertain $\rightarrow$ \texttt{False}'' policy. Although we refer to a \emph{certainty knob}, current VLMs behave more like a \emph{certainty switch}, where we cannot finely tune confidence and certainty, so the prompt enforces a conservative rule: if the model is not fully confident, answer \texttt{False}. Because the benchmark is small and only has 150 items, we evaluate each item eight times independently and aggregate the results, with confidence interval being reported. As this task appears to exceed CogVLM’s current capability, where its prediction is indistinguishable from random choice, we omit its results. As shown in Table~\ref{tab:blink_results}, the binary verification with the ``uncertain $\rightarrow$ \texttt{False}'' policy introduces a clear improvement over single-shot MCQ; the normal binary verification may introduce a slight improvement; while 3$\times$ MCQ majority voting degrades accuracy. It is known that majority voting leads to lower accuracy than single-shot when voters are below chance and/or when errors are positively correlated.  

\subsection{Discussion}
\label{sec:exp-discussion}

Across all five tasks, two consistent lessons emerge. First, quantization is critical. Moving from an open-ended prompt to a discrete MCQ formulation produces a dramatic jump in accuracy. The core reason is that, without quantization, the VLM may infer coordinates, relations, or fine-grained differences over a continuous image. After quantization, the model may only need to choose among a small number of well-defined candidates (e.g., grid cells and bounding boxes). This reduces the visual search space and aligns the language with a symbolic interface. The practical takeaway is simple: \emph{if a vision question can be quantized into an explicit shortlist of candidate states, that should be done first}. 

Second, binarization consistently provides an additional gain over MCQ. There are three main drivers. One is that answering a True/False question about a single candidate is usually easier than handling many candidates all at once. This is especially visible in REC, where MCQ requires showing multiple overlaid bounding boxes at once, which is visually noisy and induces cross-box interference, while the binary check inspects one highlighted box at a time, which is a simpler perceptual. The other driver is the deterministic resolution procedure, which plays a role similar to error detection in channel coding. Additionally, for challenging tasks, the certainty knob can be enabled to yield non-negative net gain. The lesson is that \emph{our binarization scheme is an efficient alternative to majority voting when repeated queries are allowed within the latency budget}.

\section{System}
\label{sec:system}

\begin{figure*}[t]
  \centering
  \begin{subfigure}[t]{0.44\textwidth}
    \centering
    \includegraphics[height=4.2cm,keepaspectratio]{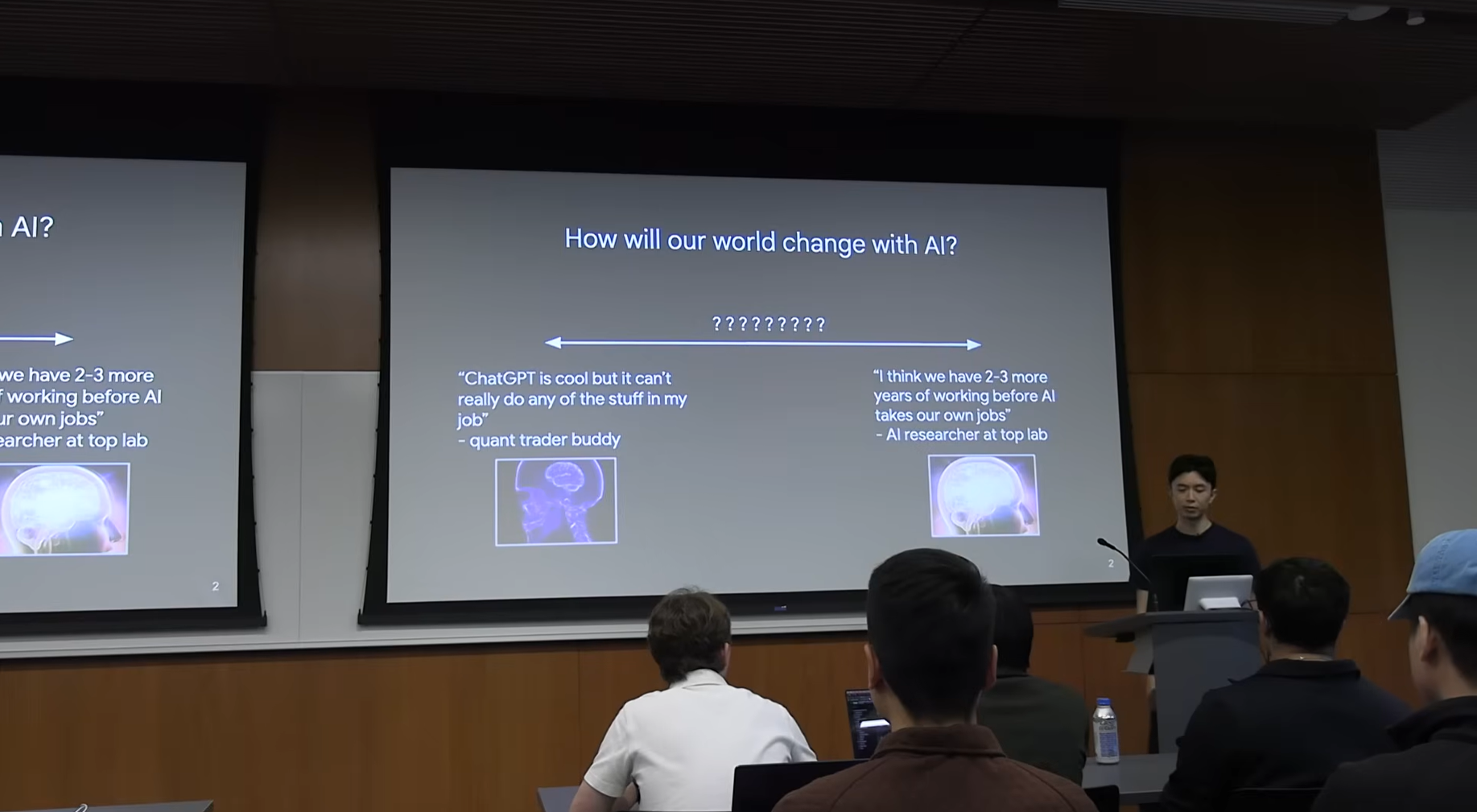}
    \caption{\textbf{Single-camera input.} A wide shot from a single fixed camera during a talk.}
    \label{fig:system:singlecam}
  \end{subfigure}
  \hspace{0.04\textwidth}
  \begin{subfigure}[t]{0.44\textwidth}
    \centering
    \includegraphics[height=4.2cm,keepaspectratio]{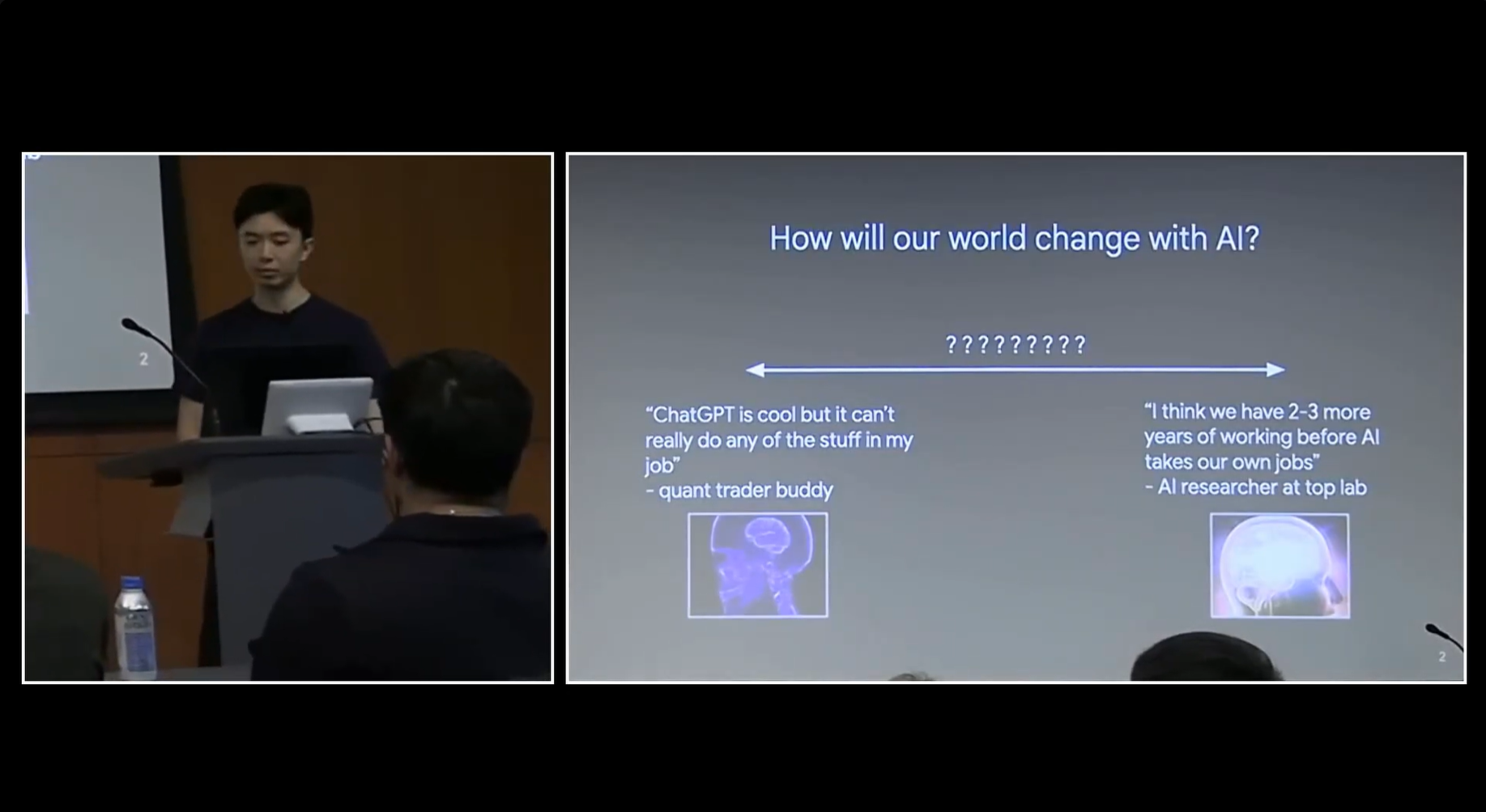}
    \caption{\textbf{System output.} Our system identifies the speaker and the main screen, applies virtual camera movement to maintain speaker and refines the screen region for a professional composited layout.}
    \label{fig:system:layout}
  \end{subfigure}
  \caption{\textbf{Real system screenshots.} Example from a talk: (a) the original single-camera scene; (b) the refined, professionally directed layout produced by our video system.}
  \label{fig:system:talk}
  \vspace{-6pt}
\end{figure*}

\begin{figure*}[t]
  \centering
  \includegraphics[width=0.96\textwidth]{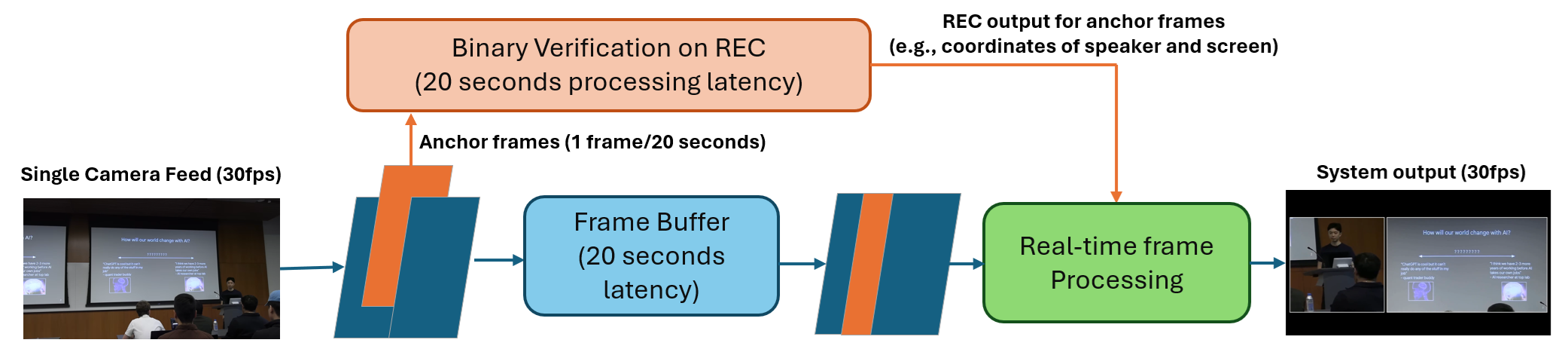}
  \caption{\textbf{Video system architecture:} End-to-end pipeline for live-streaming operation.}
  \label{fig:system:arch}
  \vspace{-6pt}
\end{figure*}

We integrate binary verification into a production video processing and editing system that reorganizes a single wide camera feed into structured, multi-view outputs (e.g., virtual camera movement and composited layouts). Active speaker (and avoid confusing them with nearby audience members) to drive speaker-focused virtual camera moves and picture-in-picture layouts, as shown in Figure~\ref{fig:system:talk}; for basketball, it must identify and track the ball in-play despite distractors such as spare balls on racks, balls held by players or referees, and ball-like patterns in the background; and for weddings, it must consistently distinguish the bride from visually similar guests (e.g., other white dresses) to produce stable, bride-centered framing and highlight moments.


Figure~\ref{fig:system:arch} shows our two-speed video pipeline that turns a single wide 30\,fps camera feed into a professionally composed output. The key is to decouple \emph{throughput} from \emph{semantic reasoning}: a frame buffer provides a modest end-to-end delay budget (e.g., $\sim$20\,s), within which we run the VLM-based \emph{binary verification} module only on periodic \emph{anchor frames} (e.g., 1 frame / 20\,s) to solve challenging image comprehension tasks such as REC and output reliable targets (e.g., speaker and screen regions). The real-time module then maintains 30\,fps by propagating these anchor-frame decisions to intermediate frames via fast tracking/association and lightweight per-frame processing (e.g., refinement, smoothing, virtual camera motion, compositing, and selective overlays/redaction). This design is practical for live streaming because many event scenarios can tolerate tens of seconds of delay (and often up to $\sim$1 minute) while preserving smooth playback; as VLM processing becomes faster, the required buffering and overall latency decrease proportionally. More broadly, because real-world scenes and queries are highly variable, we favor a training-free, zero-shot workflow that composes off-the-shelf VLMs with widely used vision models (e.g., YOLO and SAM) rather than training task-specific models, mirroring the LLM-agent paradigm of building task-specialized controllers on top of common foundation models.

Because verification is performed only on a small number of anchor frames (a few images per minute), token and compute costs remain modest; with falling token prices, this cadence is cost-effective in practice.

\section{Conclusion}

We propose a training-free binary verification workflow for zero-shot vision tasks. Across REC, spatial reasoning, and BLINK-Jigsaw, the workflow yields consistent accuracy gains. Our results also suggest that the two stages of the workflow play different roles. The \textbf{quantization} step is best viewed as a practical guideline: it converts an open-ended problem into a structured candidate set, but its exact form may vary across tasks. In contrast, the \textbf{binarization} step is much more unified. Once candidates are formed, the same binary verification procedure transfers naturally across domains, suggesting that this is the core inference mechanism of the framework.

Another important finding is the comparison with \textbf{majority voting}. Binary verification shows significant gains under comparable settings, indicating that its benefit is not simply due to repeated querying. This suggests that decomposing prediction into binary checks may provide a more reliable way to handle uncertainty. We believe this deserves further study, with more careful analysis of why the gain arises and when it is strongest.

Beyond benchmark results, we also present an end-to-end video processing system that integrates the REC workflow into a live-streaming pipeline. The system is already in production, showing that such a hybrid model architecture can be both simple and practical. More broadly, this suggests that combining a general model with a lightweight verification stage may be an effective way to improve reliability without retraining.
\clearpage

{\small
\bibliographystyle{ieee_fullname}
\bibliography{egbib}
}

\end{document}